\documentclass[10pt,twocolumn,letterpaper]{article}

\usepackage[pagenumbers]{iccv} 

\definecolor{iccvblue}{rgb}{0.21,0.49,0.74}
\usepackage[pagebackref,breaklinks,colorlinks,allcolors=iccvblue]{hyperref}
\usepackage{multirow}
\usepackage{multicol}
\usepackage{makecell}
\usepackage{bbding}

\def\paperID{7960} 
\def\confName{ICCV}
\def\confYear{2025}

\title{ReconDreamer++: Harmonizing Generative and Reconstructive Models\\for Driving Scene Representation}

\author{
~~~~~~Guosheng Zhao\footnotemark[1]~\textsuperscript{\rm \ 1, 2}
~~~~~~Xiaofeng Wang\footnotemark[1]~\textsuperscript{\rm \ 1, 2}
~~~~~~Chaojun Ni\footnotemark[1]~\textsuperscript{\rm \ 1, 3}
~~~~~~Zheng Zhu\footnotemark[1]~\textsuperscript{\rm \ 1}\textsuperscript{\Envelope} \\
~~~~~~Wenkang Qin\textsuperscript{\rm 1}
~~~~~~Guan Huang\textsuperscript{\rm 1} 
~~~~Xingang Wang\textsuperscript{\rm 2}\textsuperscript{\Envelope}
\\
\textsuperscript{\rm 1}GigaAI
~ ~ \textsuperscript{\rm 2}Institute of Automation, Chinese Academy of Sciences
~ ~ \textsuperscript{\rm 3}Peking University
\\
\small{Project Page: \url{https://recondreamer-plus.github.io}}
}

\begin{document}

\twocolumn[{%
\maketitle
\vspace{-2em}
\begin{center}
\centering
\resizebox{\linewidth}{!}{
\includegraphics{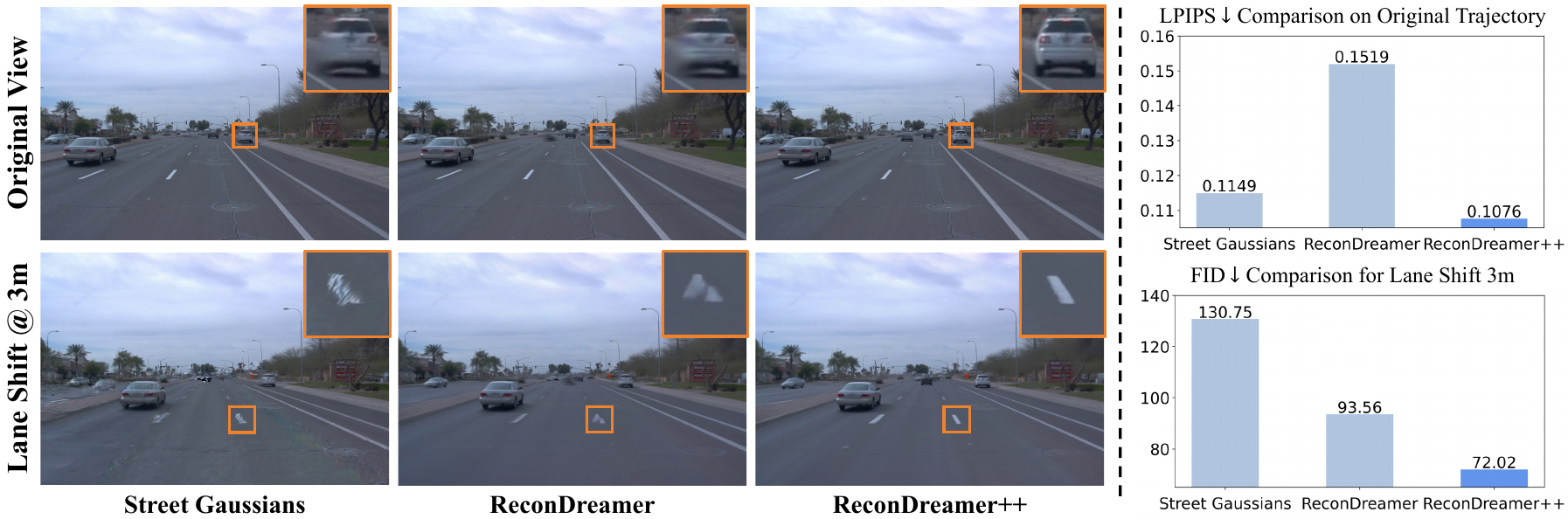}
}
\captionof{figure}{
Comparison of ReconDreamer++ with SOTA methods, Street Gaussians \cite{streetgaussian} and ReconDreamer \cite{recondreamer}, on original and novel trajectories. \textbf{Left}: ReconDreamer++ demonstrates superior rendering performance for both vehicle foregrounds and road surfaces compared to existing SOTA methods. \textbf{Right}: ReconDreamer++ significantly improves performance on novel trajectories while maintaining high rendering quality on the original trajectory.
}
\label{fig:1}
\end{center}}]

\begin{abstract}

Combining reconstruction models with generative models has emerged as a promising paradigm for closed-loop simulation in autonomous driving. For example, ReconDreamer has demonstrated remarkable success in rendering large-scale maneuvers. However, a significant gap remains between the generated data and real-world sensor observations, particularly in terms of fidelity for structured elements, such as the ground surface. To address these challenges, we propose ReconDreamer++, an enhanced framework that significantly improves the overall rendering quality by mitigating the domain gap and refining the representation of the ground surface.
Specifically, ReconDreamer++ introduces the Novel Trajectory Deformable Network (NTDNet), which leverages learnable spatial deformation mechanisms to bridge the domain gap between synthesized novel views and original sensor observations. 
Moreover, for structured elements such as the ground surface, we preserve geometric prior knowledge in 3D Gaussians, and
the optimization process focuses on refining appearance attributes while preserving the underlying geometric structure. 
Experimental evaluations conducted on multiple datasets (Waymo, nuScenes, PandaSet, and EUVS) confirm the superior performance of ReconDreamer++. Specifically, on Waymo, ReconDreamer++ achieves performance comparable to Street Gaussians for the original trajectory while significantly outperforming ReconDreamer on novel trajectories. In particular, it achieves substantial improvements, including a 6.1\% increase in NTA-IoU, a 23. 0\% improvement in FID, and a remarkable 4.5\% gain in the ground surface metric NTL-IoU, highlighting its effectiveness in accurately reconstructing structured elements such as the road surface.


\end{abstract}

\section{Introduction}
Robust driving scene representation is of paramount importance for autonomous driving simulation systems. With advances in Neural Radiance Fields (NeRF) \cite{nerf, emernerf, streetsurf, Neurad} and 3D Gaussian Splatting (3DGS) \cite{3dgs, 4dgs, deformablegs, pvg, s3gaussian, streetgaussian}, substantial progress has been achieved in reconstructing autonomous driving scenes along recorded trajectories. By incorporating generative priors \cite{sd,sdxl,svd, drivedreamer, drivedreamer2, vista}, the performance of scene representation \cite{sgd, drivedreamer4d, recondreamer, drivex, streetcrafter} in novel trajectories has been significantly improved. These developments have made the realization of closed-loop simulation for autonomous driving systems increasingly feasible.



Specifically, DriveDreamer4D \cite{drivedreamer4d} introduces a world model \cite{drivedreamer2} to directly generate videos for novel trajectories, which are then used in mixed training with 4DGS models. This significantly improves the quality of the reconstruction model under novel trajectories. Furthermore, ReconDreamer \cite{recondreamer} leverages a crafted world model for video restoration, enhancing the spatial consistency of reconstructed outputs under large maneuvers. However, existing methods that mix generated data with real-world sensor observations to train the same set of Gaussian parameters often overlook the inherent gap between these two data sources. This oversight has two critical consequences. First, it results in significant quality degradation on the original trajectory, as illustrated in Fig.~\ref{fig:1}, where ReconDreamer, although it performs well under a lane shift of 3 meters, exhibits a noticeable performance decline on the original trajectory. Second, this neglect exacerbates the low-quality rendering of structured elements within the scene, such as the ground surface. A key challenge lies in the fact that the ground surface is typically parallel to the camera's optical axis, presenting significant difficulties for traditional reconstruction methods and leading to poor robustness in modeling the ground surface. As a result, inconsistencies between generated and real-world data are further exacerbated in this region, causing suboptimal modeling quality and increased errors.

To address the aforementioned challenges, we propose ReconDreamer++, an innovative approach that significantly improves the overall rendering quality by mitigating the domain gap and refining the representation of the ground surface. Specifically, we introduce the Novel Trajectory Deformable Network (NTDNet) to systematically bridge the inherent domain gap between generated data and recorded observations. This network is specifically designed to model and learn spatial deformations, enabling a coherent mapping between synthesized and real-world data. Moreover, we independently model the road surface by incorporating 3D geometric priors into the ground Gaussian model, with an emphasis on optimizing appearance attributes while preserving the underlying geometric structure. As depicted in Fig.~\ref{fig:1}, experimental results demonstrate the superior performance of ReconDreamer++ along original and novel trajectories. Furthermore, the exceptional performance across four widely-used datasets—Waymo \cite{waymo}, nuScenes \cite{nuscenes}, PandaSet \cite{pandaset}, and EUVS \cite{euvs}—demonstrates the robustness of ReconDreamer++. Specifically, on the Waymo dataset, ReconDreamer++ achieves significant improvements over ReconDreamer, with increases of 6.1\% in NTA-IoU, 4.5\% in NTL-IoU, and 23.0\% in FID.

The main contributions of this paper are summarized as follows:

\begin{itemize}
    \item We propose ReconDreamer++, an innovative approach that significantly improves the overall rendering quality by mitigating the domain gap and refining the representation of the ground surface.
    \item To bridge the domain gap between generated and recorded data, we introduce the NTDNet, designed to learn spatial deformations between synthesized data and real-world observations. Additionally, the ground surface is modeled independently by incorporating 3D geometric priors into the ground Gaussian model. The optimization process focuses on appearance while preserving geometric fidelity, effectively alleviating the issue of cumulative errors between reconstruction and generative models.

    \item Experimental results demonstrate the superior performance of ReconDreamer++ across multiple popular datasets (Waymo \cite{waymo}, nuScenes \cite{nuscenes}, PandaSet \cite{pandaset}, EUVS \cite{euvs}). Specifically, in Waymo, ReconDreamer++ outperforms Street Gaussians by 6.4\% in the LPIPS metric on the original trajectory. On novel trajectories, it surpasses ReconDreamer by 6.1\% in NTA-IoU, 4.5\% in NTL-IoU, and 23.0\% in FID.
\end{itemize}

\begin{figure*}[!t]
\centering
\includegraphics[width=\textwidth]{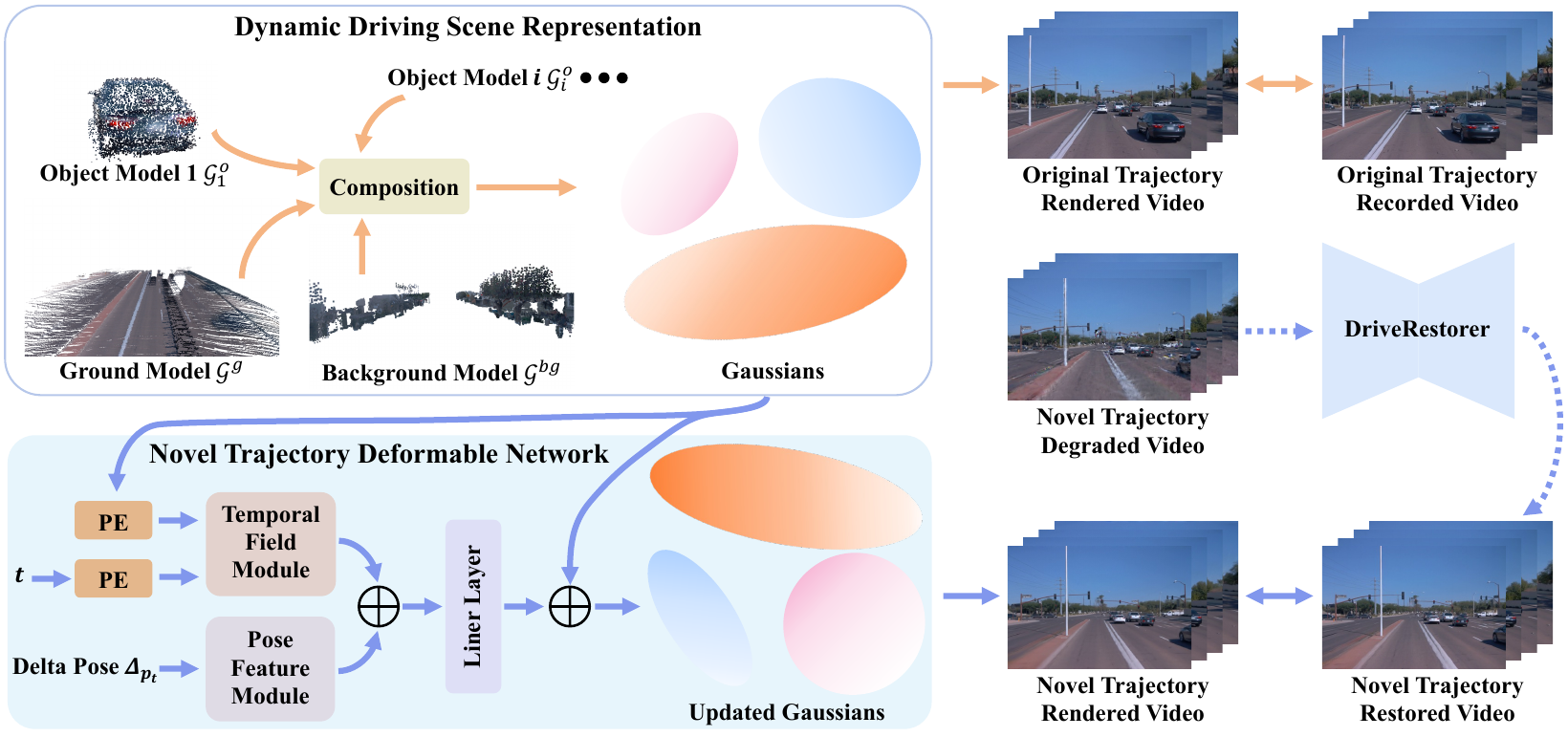}
\caption{The overall framework of ReconDreamer++. The driving scene is decomposed into three components: the ground surface, non-ground static background, and dynamic objects. For camera poses on the original trajectory, rendering is performed directly by skipping NTDNet. For camera poses on novel trajectories, the Gaussian parameters are refined through NTDNet to perform rendering.}
\label{fig_framework}
\end{figure*}

\section{Related Work}

\subsection{Driving Scene Reconstruction} 

NeRF and 3D Gaussian Splatting (3DGS) have become two of the most influential paradigms for 3D scene representation. NeRF models \cite{nerf,mipnerf,zipnerf,ngp} leverage multi-layer perceptron (MLP) networks to represent continuous volumetric scenes, enabling highly detailed reconstructions with exceptional rendering quality. On the other hand, 3DGS \cite{3dgs,mipgs} adopts a different approach by defining anisotropic Gaussians in the 3D space, using adaptive density control to produce high-quality renderings from the input of a sparse point cloud. These techniques have been adapted and extended for autonomous driving scenarios \cite{unisim, emernerf, neo360, lu2023urban, urbannerf, blocknerf, streetsurf, drivinggaussian, pvg, s3gaussian, hogaussian}, where the dynamic and complex nature of driving environments presents unique challenges. Street Gaussians \cite{streetgaussian} achieves remarkable results by decoupling dynamic scenes into foreground moving objects and static background. OmniRe \cite{omnire} further refines the dynamic foreground by distinguishing between rigid vehicles and non-rigid elements such as pedestrians and cyclists, thereby significantly enhancing the quality of dynamic driving scene reconstruction. Despite these advancements, significant limitations remain. Existing methods \cite{attal2023hyperreel,kplane,li2021neural,lin2022efficient,hypernerf,nerfplayer,ucnerf,panopticnerf,ost2021neural,Neurad,mars,snerf,khan2024autosplat, zhou2024hugsim} are highly dependent on the density and distribution of the input sensor data. When applied to scenarios that deviate from the distribution of training data, such as lane shifts, their performance tends to degrade significantly. 


\subsection{Modeling Driving Scene with Diffusion Prior}
Recent advancements in image and video generation models \cite{sd,sdxl,svd, hong2022cogvideo,yang2024cogvideox, egovid, zhu2024sora} have paved the way for a new paradigm in enhancing the representation of driving scenes by leveraging generative priors \cite{sgd,freesim,freevs,han2024ggs,vegs}. DriveDreamer4D \cite{drivedreamer4d} pioneers the use of a world model \cite{drivedreamer,drivedreamer2} to directly generate novel trajectory videos for the purpose of jointly training reconstruction models. ReconDreamer \cite{recondreamer} focuses on leveraging video generation models to refine the rendering results of novel viewpoints. By enhancing the spatial consistency between the generated outputs and the recorded sensor data, ReconDreamer achieves a significant improvement in performance when reconstructing scenes from previously unseen perspectives. These advancements collectively underscore the potential of generative approaches in addressing the challenges of novel view synthesis in autonomous driving scenarios. However, these approaches often overlook the inherent domain gap between the generated data and the original sensor observations. This neglect not only results in degraded rendering quality for the original trajectory but also amplifies the low-quality rendering of structured elements within the scene, such as the ground surface.

\section{Method}

\subsection{Preliminary: ReconDreamer}
ReconDreamer \cite{recondreamer} proposes the incremental integration of world model knowledge, significantly enhancing the model's performance for large maneuvers. The training process of ReconDreamer is divided into two stages. In the first stage, a partially trained reconstruction model is utilized to render videos along the original trajectory. This process naturally introduces ghosting artifacts or blurriness, a direct consequence of the model's underfitting. These degraded outputs are then paired with recorded sensor data to form training pairs, which are used to optimize the video restoration model, referred to as DriveRestorer \cite{recondreamer}. In the second stage, degraded videos of novel trajectories are first rendered, and the fine-tuned DriveRestorer is subsequently employed to refine these rendering results, enhancing both their quality and spatial consistency. These restored outputs, combined with the original video data, are then integrated into the training pipeline to further enhance the performance and robustness of the reconstruction model. This two-stage approach ensures that ReconDreamer achieves superior performance across the novel trajectories.

\subsection{Overview of ReconDreamer++}

The general pipeline of ReconDreamer++ is illustrated in Fig.~\ref{fig_framework}. First, the driving scene is decomposed into three components: ground surface, non-ground static background, and dynamic objects. By independently modeling the road surface, the robustness of the reconstruction model is improved, strengthening spatial consistency between generated data and recorded observations. Next, the Novel Trajectory Deformable Network (NTDNet) is introduced to systematically address the domain gap between generated and observed data. Finally, the ReconDreamer framework is adopted to restore the rendered videos along novel trajectories. The repaired results are then used to supervise and further refine the reconstruction model.

\subsection{Driving Scene Representation}
ReconDreamer \cite{recondreamer} builds upon Street Gaussians \cite{streetgaussian}, separately modeling dynamic foreground objects and static background elements while jointly optimizing them, achieving notable success. However, as illustrated in Fig.~\ref{fig:1}, this approach exhibits a lack of 3D consistency between the rendered lane markings in novel trajectories and the original data. This limitation is largely attributed to the fact that in autonomous driving scenarios, the camera's optical axis is generally parallel to the ground. Consequently, when viewed from novel perspectives, the reconstruction of the road surface often becomes suboptimal, frequently introducing streak-like or blurry artifacts. Using these rendered results, which lack 3D consistency constraints, as conditions for DriveRestorer \cite{recondreamer} can further amplify these errors, leading to a degradation in overall performance. 

To alleviate this issue, we separately model the road surface, with the entire driving scene representation consisting of three components: the ground, non-ground static background, and dynamic objects. 

\noindent{\textbf{Ground Model.}} The ground model is represented by a set of Gaussians ${\mathcal{G}}^{g}$ in the global coordinate system. Each Gaussian \cite{3dgs} is parameterized by its central position $\boldsymbol{x}$, opacity $\boldsymbol{\gamma}$, covariance $\boldsymbol{\Sigma}$, and view-dependent RGB color $\boldsymbol{c}$, controlled via spherical harmonics. For stability, each covariance matrix $\boldsymbol{\Sigma}$ is decomposed by:
\begin{equation}
    {\boldsymbol{\Sigma}} = {\boldsymbol{RSS}}^{T}{\boldsymbol{R}}^{T},
\end{equation}
where scaling matrix $\boldsymbol{S}$ and a rotation matrix $\boldsymbol{R}$ are learnable parameters, represented by scaling $\boldsymbol{s}$ and quaternion $\boldsymbol{r}$. To integrate 3D structural priors into the ground model, we employ \cite{zermas2017fast} to segment road point clouds, which are used to initialize the ground model. During training, the position $\boldsymbol{x}$ of the ground model $\mathcal{G}^g$ is fixed, and only the remaining parameters are optimized. Moreover, due to the rich availability of multi-frame fused ground point cloud data, we directly fix the number of Gaussians in the ground model, thereby avoiding any modifications to the count of Gaussian spheres. This explicit incorporation of 3D LiDAR priors significantly reduces the search space for the reconstruction model while enhancing its generalization capabilities.

\noindent{\textbf{Non-ground Background Model.}} In the world coordinate system, the non-ground background model $\mathcal{G}^{bg}$ is initialized using non-ground point cloud data, excluding moving objects. To further enhance the expressive capacity of the model, additional points are randomly generated for initialization, following the strategy outlined in \cite{pvg}. In contrast to the ground model, all parameters of the background model $\mathcal{G}^{bg}$ are fully optimized during training. 

\noindent{\textbf{Object Model.}} The dynamic object models are defined by $\{\mathcal{G}^{o}_{i}, 1\le i\le N\}$ in the local object coordinate system, where $N$ denotes the number of dynamic objects in a specific scenario. Since the object models are defined in a different coordinate system than the ground and non-ground background models, they need to be transformed into the world coordinate system during rendering based on the vehicle's pose. Suppose $\boldsymbol{x}_o$ and $\boldsymbol{q}_o$ denote the position and rotation quaternion of the Gaussian in the local object coordinate system, respectively. The pose of the dynamic object at timestep $t$ is represented by the rotation matrix $R_t$ and the translation vector $T_t$. Then, the position $\boldsymbol{x}_{w}$ and rotation quaternion $\boldsymbol{q}_w$ of the corresponding Gaussian in the world coordinate system can be obtained as follows:
\begin{equation}
    \boldsymbol{x}_w = R_t\boldsymbol{x}_o+T_t, \quad\quad \boldsymbol{q}_w = \text{ROT}(R_t,\boldsymbol{q}_o),
\end{equation}
where $\text{ROT}(\cdot)$ denotes the operation of rotating the quaternion by the rotation matrix. Additionally, multi-frame fused object point clouds are used to initialize the dynamic object models.

\begin{figure*}[!t]
\centering
\includegraphics[width=\textwidth]{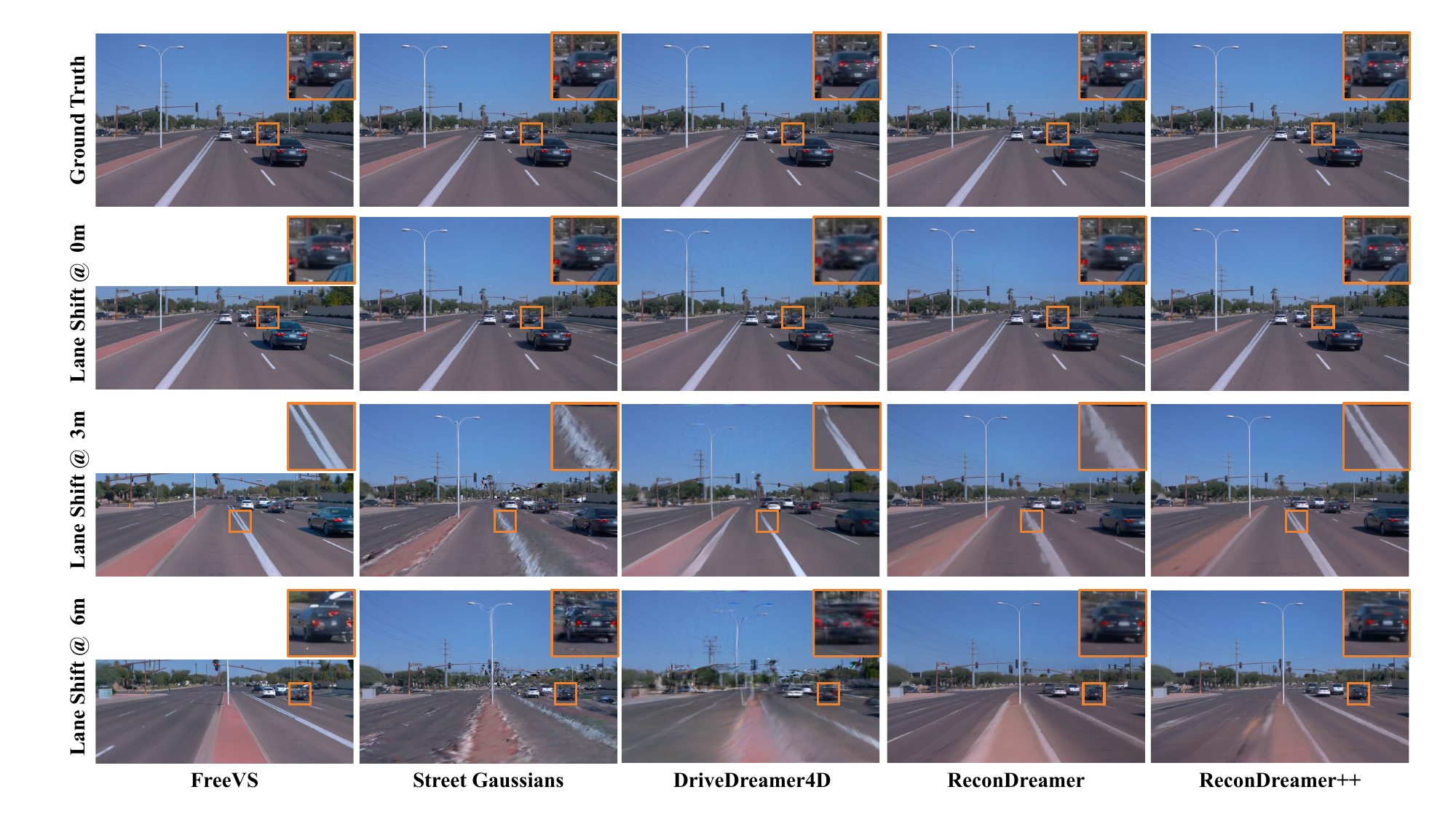}
\caption{Qualitative comparisons of different trajectory renderings on Waymo \cite{waymo}. The orange boxes highlight that ReconDreamer++ significantly enhances the rendering quality across various SOTA methods \cite{freevs,streetgaussian,drivedreamer4d,recondreamer}.}
\label{fig_waymo}
\end{figure*}

\subsection{Novel Trajectory Deformable Network}
Existing methods \cite{drivedreamer4d,recondreamer,drivex,streetcrafter} improve performance along novel trajectories by leveraging generative priors. However, these approaches often experience a noticeable decline in performance when applied to the original trajectories. To effectively bridge the domain gap between generated results and original observations, the Novel Trajectory Deformable Network (NTDNet) is proposed. As illustrated in Fig.~\ref{fig_framework}, NTDNet comprises a pose feature module $\mathcal{F}_{\phi}$ and a temporal field module $\mathcal{F}_{\theta}$. Given the original trajectory of the ego car $ \mathcal{T}^{\text{ori}} = \{p^{\text{ori}}_t, 1 \leq t \leq F\} $ and the novel trajectory $ \mathcal{T}^{\text{novel}} = \{p^{\text{novel}}_t, 1 \leq t \leq F\} $, the corresponding delta pose $\Delta p_t$ can be calculated by:
\begin{equation}
    \Delta p_t = \frac{p^{\text{novel}}_t - p^{\text{ori}}_t}{L},
\end{equation}
where $L$ refers to a hyperparameter used to normalize the delta pose. This computed delta pose $\Delta p_t$ serves as the input for the pose feature module $\mathcal{F}_\phi$. For the temporal field network $\mathcal{F}_{\theta}$, its inputs are the timestep $t$ and the Gaussian parameters $g$. Then we can obtain the delta Gaussian parameters $\Delta g$:
\begin{equation}
    \Delta g = \mathcal{F}_{out}(\mathcal{F}_{\phi}(\Delta p_t)+\mathcal{F}_{\theta}(\text{PE}(g),\text{PE}(t)))),
\end{equation}
where $\text{PE}(\cdot)$ represents the positional encoding \cite{deformablegs}, and $\mathcal{F}_{out}$ refers to the output linear layer. Additionally, both $\mathcal{F}_{\phi}$ and $\mathcal{F}_{\theta}$ are MLP layers. Finally, the updated Gaussian parameters $g'=g+\Delta g$ are used to render the video along the novel trajectory. Notably, the NTDNet is applied exclusively to the novel trajectory. In contrast, for the original trajectory, rendering is performed directly using the Gaussian parameters $ g $ without any pose adjustments.

\subsection{Optimization}
Different from ReconDreamer \cite{recondreamer}, we introduce depth supervision for synthesized data. Due to the domain gap between generated and recorded data, the direct use of point cloud re-projection of a single frame may result in performance degradation \cite{drivedreamer4d}. To mitigate this issue, we leverage multi-frame fused static point clouds to provide depth constraints for novel viewpoints. Specifically, we first fuse the point clouds from all frames to obtain $Pts_{fuse}$:
\begin{equation}
    Pts_{fuse} = \bigcup_{t=1}^F\{p\in Pts_t\ |\ \text{IsStatic}(p) \} ,
\end{equation}
where $Pts_t$ represents the point cloud of the $t$-th frame in the global coordinate system, and $\text{IsStatic}(\cdot)$ denotes a function that determines whether a point cloud is static. 

Then using the camera pose $C_{novel}$ of the novel viewpoint, the coarse depth map for the novel trajectory can be obtained. Finally, by leveraging the dynamic object mask $M_{novel}$ in the novel viewpoint, a static background depth map $D_{novel}$ can be obtained by filtering out the depth of dynamic objects as follows:
\begin{equation}
    D_{novel} = \text{Proj}(Pts_{fuse},C_{novel})\odot (1-M_{novel}),
\end{equation}
where $\text{Proj}(\cdot)$ denotes the camera projection transformation, and $\odot$ represents the element-wise multiplication. 

The overall training loss is:
\begin{equation}
    \mathcal{L} = \lambda_{1}\mathcal{L}_{\text{RGB}}+\lambda_{2}\mathcal{L}_{\text{SSIM}}+ \lambda_{3}\mathcal{L}_{\text{Depth}},
\end{equation}
where $\mathcal{L}_{\text{RGB}}$, $\mathcal{L}_{\text{SSIM}}$, $\mathcal{L}_{\text{Depth}}$ are reconstruction losses typically used in the Gaussian Splatting optimization \cite{streetgaussian}, and $\lambda_{1}$, $\lambda_{2}$ and $\lambda_{3}$ are loss weights.

\begin{table*}[t]
\centering
\resizebox{1\linewidth}{!}{
\begin{tabular}{@{}m{4.6cm}cccccccccc@{}}
\toprule
\multirow{2}{*}{Method} & \multicolumn{3}{c}{Lane Shift @ 0m} & \multicolumn{3}{c}{Lane Shift @ 3m} & \multicolumn{3}{c}{Lane Shift @ 6m} \\ 
\cmidrule(lr){2-4} \cmidrule(lr){5-7} \cmidrule(lr){8-10} 
 &PSNR $\uparrow$ & SSIM $\uparrow$ & LPIPS $\downarrow$ & NTA-IoU $\uparrow$ & NTL-IoU $\uparrow$ & FID$\downarrow$ & NTA-IoU $\uparrow$ & NTL-IoU $\uparrow$ & FID$\downarrow$  \\ 
 \midrule
Street Gaussians \cite{streetgaussian} & \textbf{36.50}&0.9570 &0.1149 & 0.498 & 53.19  & 130.75 & 0.374 & 49.27 & 213.04 \\
FreeVS \cite{freevs} &26.59 &0.8057 &0.1325 & 0.505 & 56.84  & 104.23 &0.465 &55.37  &121.44 \\
DriveDreamer4D with PVG \cite{drivedreamer4d}&34.37 &0.9447 &0.1321 &0.457  &53.30  &113.45 &0.159 &50.09 & 261.81 \\
ReconDreamer \cite{recondreamer} & 34.31 & 0.9431 & 0.1519 &0.539 & 54.58 & 93.56  & 0.467 & 52.58 & 149.19 \\
\midrule
ReconDreamer++ & 36.29 & \textbf{0.9573}&\textbf{0.1076} &\textbf{0.572} &\textbf{57.06} &\textbf{72.02} &\textbf{0.489} &\textbf{56.57} &\textbf{111.92} \\
\bottomrule
\end{tabular}}
\caption{Comparison of different lane shifts on the Waymo dataset with various methods \cite{streetgaussian,freevs,drivedreamer4d,recondreamer}.}
\label{tab:waymo}
\end{table*}

\begin{table*}[t]
\centering
\resizebox{1\linewidth}{!}{
\begin{tabular}{@{}m{4.6cm}cccccccccc@{}}
\toprule
\multirow{2}{*}{Method} & \multicolumn{3}{c}{Lane Shift @ 0m} & \multicolumn{3}{c}{Lane Shift @ 3m} & \multicolumn{3}{c}{Lane Shift @ 6m} \\ 
\cmidrule(lr){2-4} \cmidrule(lr){5-7} \cmidrule(lr){8-10} 
 &PSNR $\uparrow$ & SSIM $\uparrow$ & LPIPS $\downarrow$ & NTA-IoU $\uparrow$ & NTL-IoU $\uparrow$ & FID$\downarrow$ & NTA-IoU $\uparrow$ & NTL-IoU $\uparrow$ & FID$\downarrow$  \\ 
 \midrule
Street Gaussians \cite{streetgaussian} &\textbf{34.91} &0.9528 &0.1185 & 0.219 &48.72 &149.50  & 0.132& 46.57 &247.73  \\
ReconDreamer \cite{recondreamer} &33.85 &0.9497 &0.1235 &0.325 &49.72 &125.43 &0.246 &  47.08&197.52 \\
\midrule  
ReconDreamer++ &34.69 &\textbf{0.9533} &\textbf{0.1161} &\textbf{0.365} &\textbf{51.33} &\textbf{118.27} &\textbf{0.268} & \textbf{49.26} & \textbf{187.51}\\
\bottomrule
\end{tabular}}
\caption{Comparison of different lane shifts on the nuScenes dataset with various methods \cite{streetgaussian,recondreamer}.}
\label{tab:nuscenes}

\end{table*}

\section{Experiments}
In this section, we present our experimental setup, which encompasses the datasets, baselines, implementation details, and evaluation metrics. Subsequently, both quantitative and qualitative results are provided to demonstrate the superior performance of the proposed ReconDreamer++. Finally, we perform experiments to investigate the effectiveness of the depth loss, ground model, and NTDNet. 

\subsection{Experiment Setup}
\noindent
\textbf{Dataset.} We conduct our experiments on various datasets, including Waymo \cite{waymo}, nuScenes \cite{nuscenes}, PandaSet \cite{pandaset} and EUVS \cite{euvs}. For Waymo, we follow the configuration of ReconDreamer \cite{recondreamer}. For nuScenes, we select eight scenes rich in dynamic objects and complex lane structures to test large maneuver capabilities. For PandaSet, we adopt the ten scenes used in Unisim \cite{unisim} and NeuRAD \cite{Neurad} for fair comparison. For EUVS, we select four scenes from Level 1 for evaluation. Specifically, we use only a single video segment from the training set for training and employ the test set (offset by one lane from the training set) for evaluation. These datasets enable comprehensive performance analysis across diverse driving scenarios.

\noindent
\textbf{Baselines.} Since existing methods typically conduct experiments on one or two datasets with varying scene selections, we chose different state-of-the-art (SOTA) methods for comparison across different datasets. Specifically, on Waymo, we compare with FreeVS \cite{freevs}, Street Gaussians \cite{streetgaussian}, DriveDreamer4D \cite{drivedreamer4d}, and ReconDreamer \cite{recondreamer}. For nuScenes and EUVS, we reproduce and evaluate Street Gaussians and ReconDreamer on our selected scenes. On PandaSet, we include comparisons with Street Gaussians, ReconDreamer, UniSim \cite{unisim}, and NeuRAD \cite{Neurad}. 

\noindent
\textbf{Implementation details.}  The training strategies and hyperparameters are configured to align with those of ReconDreamer \cite{recondreamer}, with a total of 50,000 training steps. Both the temporal field module and pose feature module are lightweight networks, implemented using 8 layers of MLPs with a hidden dimension of 256.


\noindent
\textbf{Metrics.} 
For the evaluation of the original trajectory, we use PSNR, SSIM, and LPIPS \cite{lpips} as metrics. For novel trajectories, we follow the evaluation protocol of DriveDreamer4D \cite{drivedreamer4d}, utilizing three key metrics: Novel Trajectory Agent IoU (NTA-IoU) to assess foreground quality, Novel Trajectory Lane IoU (NTL-IoU) to evaluate background lane markings, and Fréchet Inception Distance (FID) \cite{fid} to measure overall rendering fidelity.

\begin{figure*}[!t]
\centering
\includegraphics[width=\textwidth]{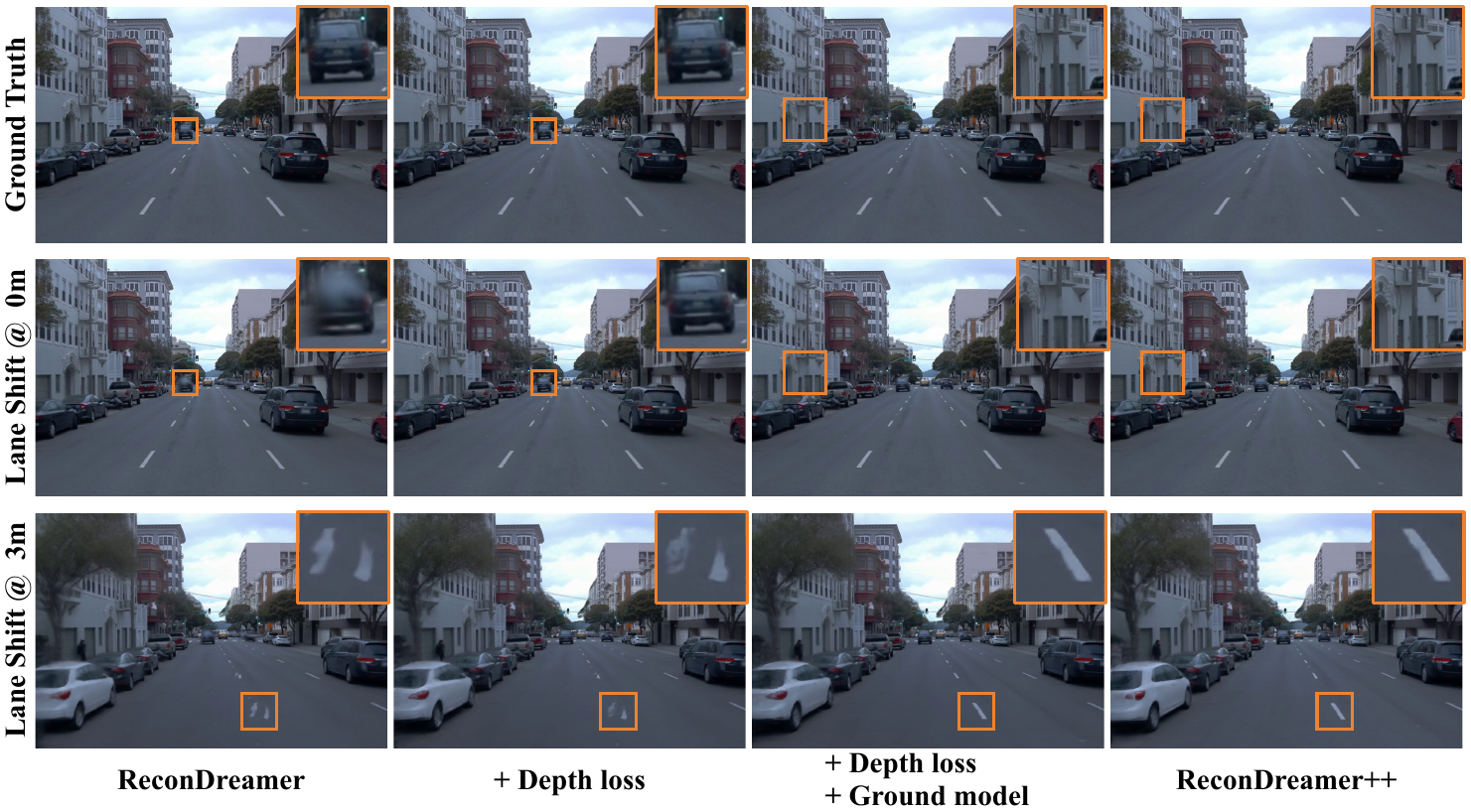}
\caption{Ablation studies on the depth loss, ground model and NTDNet. When none of these components are employed, the model corresponds to ReconDreamer. In contrast, when all of these components are integrated, the model represents ReconDreamer++. This highlights the incremental contributions of each component in enhancing the overall performance.}
\label{fig_ablation}
\end{figure*}

\subsection{Main Results}
\noindent
\textbf{Waymo.} 
For Waymo dataset \cite{waymo}, we compare our method with several SOTA approaches \cite{streetgaussian, freevs, drivedreamer4d, recondreamer}. Notably, the generation results of FreeVS \cite{freevs} rely on LiDAR point clouds, which means that areas in the upper part of the image that are typically not scanned by LiDAR are ignored by this method. Therefore, when calculating metrics, we crop the corresponding regions of the ground truth images accordingly. The results shown in Tab.~\ref{tab:waymo} demonstrate that our ReconDreamer++ not only achieves performance comparable to traditional reconstruction methods like Street Gaussians \cite{streetgaussian} on the original trajectory but also improves the LPIPS metric by 6.4\%. However, FreeVS, ReconDreamer \cite{recondreamer}, and DriveDreamer4D \cite{drivedreamer4d} all exhibit significant quality degradation on the original trajectory. Additionally, on novel trajectories, ReconDreamer++ comprehensively outperforms the SOTA method ReconDreamer. Specifically, for the lane shift of 3 meters, it achieves improvements of 6.1\% in NTA-IoU, 4.5\% in NTL-IoU, and 23.0\% in FID. For the lane shift of 6 meters, it achieves gains of 4.7\% in NTA-IoU, 7.6\% in NTL-IoU, and 25.0\% in FID. Fig.~\ref{fig_waymo} demonstrates that ReconDreamer++ achieves results comparable to Street Gaussians on the original trajectory, while surpassing all SOTA methods on novel trajectories, achieving the best performance. Specifically, under a 3-meter lane shift, ReconDreamer++ maintains spatial consistency of lane markings with respect to the ground truth. In contrast, ReconDreamer fails to preserve this consistency, highlighting the superiority of ReconDreamer++ in handling challenging scenarios.

\begin{table}[t]
\centering
\resizebox{1\linewidth}{!}{
\begin{tabular}{@{}m{3cm}cccc}
\toprule
Method &No Shift & Lane 2m & Lane 3m & Vert. 1m  \\ 
\midrule
UniSim \cite{unisim} &- & 74.7 & 97.5 & -  \\
NeuRAD \cite{Neurad} & 25.0 & 72.3 & 93.9 & 76.3 \\ 
Street Gaussians \cite{streetgaussian} & 14.6 & 66.3 & 80.7  & 78.4 \\ 
ReconDreamer \cite{recondreamer} &16.1 & 65.4 & 74.9 & 67.7\\ 
\midrule
ReconDreamer++ &\textbf{13.8}&\textbf{61.9} &\textbf{71.7} &\textbf{62.4}  \\
\bottomrule
\end{tabular}}
\caption{FID scores for ego vehicle pose shifts in PandaSet.}
\label{tab:pandaset}
\end{table}

\noindent
\textbf{nuScenes.} As shown in Tab.~\ref{tab:nuscenes}, we reproduce  Street Gaussians \cite{streetgaussian} and ReconDreamer \cite{recondreamer} on nuScenes for comparison with ReconDreamer++. The results align with conclusion on Waymo \cite{waymo}, where ReconDreamer++ achieves performance comparable to Street Gaussians on the original trajectory. Specifically, the metrics are as follows: PSNR of 34.69 versus 34.91, SSIM of 0.9533 versus 0.9528, and LPIPS of 0.1161 versus 0.1185. For novel trajectories with a 3-meter lateral shift, ReconDreamer++ shows noteworthy improvements over ReconDreamer, achieving NTA-IoU of 0.365 (a 12.3\% improvement), NTL-IoU of 51.33 (a 3.2\% improvement), and FID of 118.27 (a 5.7\% improvement). Similarly, for trajectories with a 6-meter lateral shift, ReconDreamer++ attains NTA-IoU of 0.268 (a 8.9\% improvement), NTL-IoU of 49.26 (a 4.6\% improvement), and FID of 187.51 (a 5.1\% improvement).



\begin{table}[t]
\centering
\resizebox{1\linewidth}{!}{
\begin{tabular}{@{}m{4.6cm}ccc@{}}
\toprule
\multirow{2}{*}{Method} & \multicolumn{3}{c}{Test Set} \\ 
\cmidrule(lr){2-4}
 &PSNR $\uparrow$ &SSIM $\uparrow$ &LPIPS $\downarrow$ \\ 
 \midrule
Street Gaussians \cite{streetgaussian}&15.08&0.6043&0.3428  \\
ReconDreamer \cite{recondreamer}&20.14&0.6914&0.3286\\
\midrule  
ReconDreamer++ &\textbf{20.52}&\textbf{0.7033}&\textbf{0.3057}\\
\bottomrule
\end{tabular}}
\caption{Comparison of various methods \cite{streetgaussian,recondreamer} on the EUVS test set from Level 1.}
\label{tab:euvs}

\end{table}

\begin{table*}[t]
\centering
\resizebox{1\linewidth}{!}{
\begin{tabular}{cccccccccccc}
\toprule
\multirow{2}{*}{Depth loss}&\multirow{2}{*}{Ground model} &\multirow{2}{*}{NTDNet} & \multicolumn{3}{c}{Lane Shift @ 0m} & \multicolumn{3}{c}{Lane Shift @ 3m} & \multicolumn{3}{c}{Lane Shift @ 6m} \\ 
\cmidrule(lr){4-6} \cmidrule(lr){7-9} \cmidrule(lr){10-12} 
 & &  &PSNR $\uparrow$ & SSIM $\uparrow$ & LPIPS $\downarrow$ & NTA-IoU $\uparrow$ & NTL-IoU $\uparrow$ & FID$\downarrow$ & NTA-IoU $\uparrow$ & NTL-IoU $\uparrow$ & FID$\downarrow$  \\ 
 \midrule
-&-&-& 34.31 & 0.9431 & 0.1519 &0.539 & 54.58 & 93.56  & 0.467 & 52.58 & 149.19  \\
\checkmark&- &-& 35.13&0.9526 &0.1207 &0.561  &55.91  &77.51  &0.473 &53.55  &124.52 \\
\checkmark&\checkmark&- &34.93 &0.9518 &0.1174 &0.566  & 56.89 &75.22 &0.484 &55.34 &121.46   \\
\checkmark&\checkmark&\checkmark & \textbf{36.29} & \textbf{0.9573}&\textbf{0.1076} &\textbf{0.572} &\textbf{57.06} &\textbf{72.02} &\textbf{0.489} &\textbf{56.57} &\textbf{111.92} \\
\bottomrule
\end{tabular}}
\caption{Ablation study of the depth loss, ground model and NTDNet on the Waymo dataset.}
\label{tab:ablation}
\end{table*}

\noindent
\textbf{PandaSet.} For PandaSet \cite{pandaset}, we conduct a comprehensive comparison of our method with several SOTA approaches, including UniSim \cite{unisim}, NeuRAD \cite{Neurad}, Street Gaussians \cite{streetgaussian}, and ReconDreamer \cite{recondreamer}. Notably, the results for Street Gaussians and ReconDreamer are based on our own reproduction. To ensure consistency with the evaluation protocols of UniSim and NeuRAD, we focus exclusively on the FID metric for assessing performance. Specifically, we evaluate the FID scores on the original trajectory, as well as for novel trajectories with lateral shifts of 2m and 3m, and a vertical shift of 1m. As shown in Tab.~\ref{tab:pandaset}, our method achieves SOTA results, demonstrating superior performance in terms of rendering quality and generalization capabilities across all evaluated scenarios. These results highlight the effectiveness of our approach in addressing the challenges posed by complex driving scenes and diverse trajectory variations.

\noindent
\textbf{EUVS.} We reproduce Street Gaussians \cite{streetgaussian} and ReconDreamer \cite{recondreamer} using the EUVS dataset \cite{euvs} and report the performance comparison with ReconDreamer++ on the test set metrics in Tab.~\ref{tab:euvs}. As shown in Tab.~\ref{tab:euvs}, 
ReconDreamer++ demonstrates outstanding performance on the EUVS test set, achieving remarkable results across multiple evaluation metrics. Specifically, it scores 20.52 in PSNR, 0.7033 in SSIM, and 0.3057 in LPIPS, showcasing its ability to produce high-fidelity reconstructions. Compared to ReconDreamer, these results represent significant improvements of 1.9\% in PSNR, 1.7\% in SSIM, and an impressive 7.0\% in LPIPS. These enhancements highlight the effectiveness of ReconDreamer++ in capturing finer details and improving perceptual quality, marking a substantial advancement over its predecessor in handling complex driving scenarios.

\subsection{Ablation Study} 
ReconDreamer serves as our baseline method, and the improvements achieved by ReconDreamer++ can be attributed to three key aspects: the introduction of depth supervision for novel views, the separate modeling of the road surface, and the mitigation of the gap between generated and real data. To evaluate the individual contributions of these components, we conduct an ablation study on the depth loss, ground model, and NTDNet. The experimental results presented in Tab.~\ref{tab:ablation} and Fig.~\ref{fig_ablation} provide valuable insight into the contributions of each component. 

\noindent
\textbf{Depth loss.} The introduction of depth loss specifically for novel trajectories strengthens the model's spatial understanding, resulting in notable improvements in the overall performance of the reconstruction process. As illustrated in Fig.~\ref{fig_ablation}, the introduction of depth supervision for novel viewpoints significantly enhances the model's ability to accurately model vehicles, resulting in sharper and more detailed reconstructions. This improvement directly translates into better performance across all metrics reported in Tab.~\ref{tab:ablation}. 

\noindent
\textbf{Ground model.} Separate modeling of the ground surface plays a pivotal role in enhancing model performance under novel trajectories. As shown in Fig.~\ref{fig_ablation}, there is a marked improvement in the reconstruction quality for novel trajectories, particularly in preserving the geometric consistency of lane markings with respect to the ground truth. This enhancement can be credited to the independent modeling of the ground surface, which effectively reduces the cumulative error between the generative and reconstruction models. These qualitative observations are further supported by the quantitative results in Tab.~\ref{tab:ablation}. Notably, for large lateral shifts of 6 meters, the model achieves a 3.3\% improvement in NTL-IoU, highlighting the effectiveness of this enhancement in addressing challenging scenarios. 

\noindent
\textbf{NTDNet.} Both Fig.~\ref{fig_ablation} and Tab.~\ref{tab:ablation} demonstrate that the integration of NTDNet further elevates the model's performance across both original and novel trajectories. These results indicate that NTDNet not only bridges the gap between original observations and synthetic data but also unlocks the model's potential by enabling robust and high-quality reconstructions in diverse scenarios. 

In summary, the ablation study highlights the complementary roles of depth loss, the ground model, and NTDNet in improving accuracy, consistency, and generalization.

\section{Discussion and Conclusion}


Owing to the rapid advancements in generative models, existing reconstruction methods leveraging generative priors have made remarkable strides in rendering novel trajectories, showcasing significant potential for closed-loop simulation in autonomous driving. Despite these achievements, a critical limitation persists: the domain gap between generated and recorded data is frequently overlooked. This neglect can result in inaccuracies that severely undermine spatial consistency and compromise the overall quality of scene reconstruction, particularly in complex driving scenarios. To tackle this challenge, we introduce ReconDreamer++, a novel framework that integrates NTDNet and adopts a dedicated modeling approach for the ground surface. 
By explicitly addressing the domain gap and enhancing the representation of structured elements, ReconDreamer++ achieves substantial improvements in both original and novel trajectories. Extensive experiments conducted on multiple datasets (Waymo, nuScenes, PandaSet, and EUVS) highlight the effectiveness and robustness of our method across a wide range of scenarios.

{
    \small
    \bibliographystyle{ieeenat_fullname}
    \bibliography{PaperForReview}

\begin{thebibliography}{63}
\providecommand{\natexlab}[1]{#1}
\providecommand{\url}[1]{\texttt{#1}}
\expandafter\ifx\csname urlstyle\endcsname\relax
  \providecommand{\doi}[1]{doi: #1}\else
  \providecommand{\doi}{doi: \begingroup \urlstyle{rm}\Url}\fi

\bibitem[Attal et~al.(2023)Attal, Huang, Richardt, Zollhoefer, Kopf, O’Toole, and Kim]{attal2023hyperreel}
Benjamin Attal, Jia-Bin Huang, Christian Richardt, Michael Zollhoefer, Johannes Kopf, Matthew O’Toole, and Changil Kim.
\newblock Hyperreel: High-fidelity 6-dof video with ray-conditioned sampling.
\newblock In \emph{CVPR}, 2023.

\bibitem[Barron et~al.(2022)Barron, Mildenhall, Verbin, Srinivasan, and Hedman]{mipnerf}
Jonathan~T Barron, Ben Mildenhall, Dor Verbin, Pratul~P Srinivasan, and Peter Hedman.
\newblock Mip-nerf 360: Unbounded anti-aliased neural radiance fields.
\newblock In \emph{CVPR}, 2022.

\bibitem[Barron et~al.(2023)Barron, Mildenhall, Verbin, Srinivasan, and Hedman]{zipnerf}
Jonathan~T Barron, Ben Mildenhall, Dor Verbin, Pratul~P Srinivasan, and Peter Hedman.
\newblock Zip-nerf: Anti-aliased grid-based neural radiance fields.
\newblock In \emph{ICCV}, 2023.

\bibitem[Blattmann et~al.(2023)Blattmann, Dockhorn, Kulal, Mendelevitch, Kilian, Lorenz, Levi, English, Voleti, Letts, et~al.]{svd}
Andreas Blattmann, Tim Dockhorn, Sumith Kulal, Daniel Mendelevitch, Maciej Kilian, Dominik Lorenz, Yam Levi, Zion English, Vikram Voleti, Adam Letts, et~al.
\newblock Stable video diffusion: Scaling latent video diffusion models to large datasets.
\newblock \emph{arXiv preprint arXiv:2311.15127}, 2023.

\bibitem[Caesar et~al.(2020)Caesar, Bankiti, Lang, Vora, Liong, Xu, Krishnan, Pan, Baldan, and Beijbom]{nuscenes}
Holger Caesar, Varun Bankiti, Alex~H Lang, Sourabh Vora, Venice~Erin Liong, Qiang Xu, Anush Krishnan, Yu Pan, Giancarlo Baldan, and Oscar Beijbom.
\newblock nuscenes: A multimodal dataset for autonomous driving.
\newblock In \emph{CVPR}, 2020.

\bibitem[Che et~al.(2023)Che, Nguyen, Pham, and Lam]{che2023twinlitenet}
Quang-Huy Che, Dinh-Phuc Nguyen, Minh-Quan Pham, and Duc-Khai Lam.
\newblock Twinlitenet: An efficient and lightweight model for driveable area and lane segmentation in self-driving cars.
\newblock In \emph{MAPR}, 2023.

\bibitem[Chen et~al.(2023)Chen, Gu, Jiang, Zhu, and Zhang]{pvg}
Yurui Chen, Chun Gu, Junzhe Jiang, Xiatian Zhu, and Li Zhang.
\newblock Periodic vibration gaussian: Dynamic urban scene reconstruction and real-time rendering.
\newblock \emph{arXiv preprint arXiv:2311.18561}, 2023.

\bibitem[Chen et~al.(2024)Chen, Yang, Huang, Lutio, Esturo, Ivanovic, Litany, Gojcic, Fidler, Pavone, Song, and Wang]{omnire}
Ziyu Chen, Jiawei Yang, Jiahui Huang, Riccardo~de Lutio, Janick~Martinez Esturo, Boris Ivanovic, Or Litany, Zan Gojcic, Sanja Fidler, Marco Pavone, Li Song, and Yue Wang.
\newblock Omnire: Omni urban scene reconstruction.
\newblock \emph{arXiv preprint arXiv:2408.16760}, 2024.

\bibitem[Cheng et~al.(2023)Cheng, Long, Yin, Wang, Wu, Ma, Wang, Chen, and Chen]{ucnerf}
Kai Cheng, Xiaoxiao Long, Wei Yin, Jin Wang, Zhiqiang Wu, Yuexin Ma, Kaixuan Wang, Xiaozhi Chen, and Xuejin Chen.
\newblock Uc-nerf: Neural radiance field for under-calibrated multi-view cameras in autonomous driving.
\newblock \emph{arXiv preprint arXiv:2311.16945}, 2023.

\bibitem[Fan et~al.(2024)Fan, Zhang, Wang, Li, and Zhang]{freesim}
Lue Fan, Hao Zhang, Qitai Wang, Hongsheng Li, and Zhaoxiang Zhang.
\newblock Freesim: Toward free-viewpoint camera simulation in driving scenes.
\newblock \emph{arXiv preprint arXiv:2412.03566}, 2024.

\bibitem[Fridovich-Keil et~al.(2023)Fridovich-Keil, Meanti, Warburg, Recht, and Kanazawa]{kplane}
Sara Fridovich-Keil, Giacomo Meanti, Frederik~Rahb{\ae}k Warburg, Benjamin Recht, and Angjoo Kanazawa.
\newblock K-planes: Explicit radiance fields in space, time, and appearance.
\newblock In \emph{CVPR}, 2023.

\bibitem[Gao et~al.(2024)Gao, Yang, Chen, Chitta, Qiu, Geiger, Zhang, and Li]{vista}
Shenyuan Gao, Jiazhi Yang, Li Chen, Kashyap Chitta, Yihang Qiu, Andreas Geiger, Jun Zhang, and Hongyang Li.
\newblock Vista: A generalizable driving world model with high fidelity and versatile controllability.
\newblock \emph{arXiv preprint arXiv:2405.17398}, 2024.

\bibitem[Guo et~al.(2023)Guo, Deng, Li, Bai, Shi, Wang, Ding, Wang, and Li]{streetsurf}
Jianfei Guo, Nianchen Deng, Xinyang Li, Yeqi Bai, Botian Shi, Chiyu Wang, Chenjing Ding, Dongliang Wang, and Yikang Li.
\newblock Streetsurf: Extending multi-view implicit surface reconstruction to street views.
\newblock \emph{arXiv preprint arXiv:2306.04988}, 2023.

\bibitem[Han et~al.(2024{\natexlab{a}})Han, Zhou, Long, Wang, and Xiao]{han2024ggs}
Huasong Han, Kaixuan Zhou, Xiaoxiao Long, Yusen Wang, and Chunxia Xiao.
\newblock Ggs: Generalizable gaussian splatting for lane switching in autonomous driving.
\newblock \emph{arXiv preprint arXiv:2409.02382}, 2024{\natexlab{a}}.

\bibitem[Han et~al.(2024{\natexlab{b}})Han, Jia, Li, Wang, Ivanovic, You, Liu, Wang, Pavone, Feng, and Li]{euvs}
Xiangyu Han, Zhen Jia, Boyi Li, Yan Wang, Boris Ivanovic, Yurong You, Lingjie Liu, Yue Wang, Marco Pavone, Chen Feng, and Yiming Li.
\newblock Extrapolated urban view synthesis benchmark.
\newblock \emph{arXiv preprint arXiv:2412.05256}, 2024{\natexlab{b}}.

\bibitem[Heusel et~al.(2017)Heusel, Ramsauer, Unterthiner, Nessler, and Hochreiter]{fid}
Martin Heusel, Hubert Ramsauer, Thomas Unterthiner, Bernhard Nessler, and Sepp Hochreiter.
\newblock Gans trained by a two time-scale update rule converge to a local nash equilibrium.
\newblock \emph{NeurIPS}, 2017.

\bibitem[Hong et~al.(2022)Hong, Ding, Zheng, Liu, and Tang]{hong2022cogvideo}
Wenyi Hong, Ming Ding, Wendi Zheng, Xinghan Liu, and Jie Tang.
\newblock Cogvideo: Large-scale pretraining for text-to-video generation via transformers.
\newblock \emph{arXiv preprint arXiv:2205.15868}, 2022.

\bibitem[Huang et~al.(2024)Huang, Wei, Zheng, An, Lu, Zhan, Tomizuka, Keutzer, and Zhang]{s3gaussian}
Nan Huang, Xiaobao Wei, Wenzhao Zheng, Pengju An, Ming Lu, Wei Zhan, Masayoshi Tomizuka, Kurt Keutzer, and Shanghang Zhang.
\newblock $s^3$gaussian: Self-supervised street gaussians for autonomous driving.
\newblock \emph{arXiv preprint arXiv:2405.20323}, 2024.

\bibitem[Hwang et~al.(2024)Hwang, Kim, Kang, Kang, and Choo]{vegs}
Sungwon Hwang, Min-Jung Kim, Taewoong Kang, Jayeon Kang, and Jaegul Choo.
\newblock Vegs: View extrapolation of urban scenes in 3d gaussian splatting using learned priors.
\newblock In \emph{ECCV}, 2024.

\bibitem[Irshad et~al.(2023)Irshad, Zakharov, Liu, Guizilini, Kollar, Gaidon, Kira, and Ambrus]{neo360}
Muhammad~Zubair Irshad, Sergey Zakharov, Katherine Liu, Vitor Guizilini, Thomas Kollar, Adrien Gaidon, Zsolt Kira, and Rares Ambrus.
\newblock Neo 360: Neural fields for sparse view synthesis of outdoor scenes.
\newblock In \emph{ICCV}, 2023.

\bibitem[Jocher and Qiu(2024)]{yolo11_ultralytics}
Glenn Jocher and Jing Qiu.
\newblock Ultralytics yolo11, 2024.

\bibitem[Kerbl et~al.(2023)Kerbl, Kopanas, Leimk{\"u}hler, and Drettakis]{3dgs}
Bernhard Kerbl, Georgios Kopanas, Thomas Leimk{\"u}hler, and George Drettakis.
\newblock 3d gaussian splatting for real-time radiance field rendering.
\newblock \emph{ACM ToG}, 2023.

\bibitem[Khan et~al.(2024)Khan, Fazlali, Sharma, Cao, Bai, Ren, and Liu]{khan2024autosplat}
Mustafa Khan, Hamidreza Fazlali, Dhruv Sharma, Tongtong Cao, Dongfeng Bai, Yuan Ren, and Bingbing Liu.
\newblock Autosplat: Constrained gaussian splatting for autonomous driving scene reconstruction.
\newblock \emph{arXiv preprint arXiv:2407.02598}, 2024.

\bibitem[Kundu et~al.(2022)Kundu, Genova, Yin, Fathi, Pantofaru, Guibas, Tagliasacchi, Dellaert, and Funkhouser]{panopticnerf}
Abhijit Kundu, Kyle Genova, Xiaoqi Yin, Alireza Fathi, Caroline Pantofaru, Leonidas~J Guibas, Andrea Tagliasacchi, Frank Dellaert, and Thomas Funkhouser.
\newblock Panoptic neural fields: A semantic object-aware neural scene representation.
\newblock In \emph{CVPR}, 2022.

\bibitem[Li et~al.(2021)Li, Niklaus, Snavely, and Wang]{li2021neural}
Zhengqi Li, Simon Niklaus, Noah Snavely, and Oliver Wang.
\newblock Neural scene flow fields for space-time view synthesis of dynamic scenes.
\newblock In \emph{CVPR}, 2021.

\bibitem[Li et~al.(2024)Li, Zhang, Wu, Zhu, and Zhang]{hogaussian}
Zhuopeng Li, Yilin Zhang, Chenming Wu, Jianke Zhu, and Liangjun Zhang.
\newblock Ho-gaussian: Hybrid optimization of 3d gaussian splatting for urban scenes.
\newblock \emph{arXiv preprint arXiv:2403.20032}, 2024.

\bibitem[Lin et~al.(2022)Lin, Peng, Xu, Yan, Shuai, Bao, and Zhou]{lin2022efficient}
Haotong Lin, Sida Peng, Zhen Xu, Yunzhi Yan, Qing Shuai, Hujun Bao, and Xiaowei Zhou.
\newblock Efficient neural radiance fields for interactive free-viewpoint video.
\newblock In \emph{SIGGRAPH Asia}, 2022.

\bibitem[Lu et~al.(2023)Lu, Xu, Chen, Li, Lin, and Jiang]{lu2023urban}
Fan Lu, Yan Xu, Guang Chen, Hongsheng Li, Kwan-Yee Lin, and Changjun Jiang.
\newblock Urban radiance field representation with deformable neural mesh primitives.
\newblock In \emph{ICCV}, 2023.

\bibitem[Mildenhall et~al.(2021)Mildenhall, Srinivasan, Tancik, Barron, Ramamoorthi, and Ng]{nerf}
Ben Mildenhall, Pratul~P Srinivasan, Matthew Tancik, Jonathan~T Barron, Ravi Ramamoorthi, and Ren Ng.
\newblock Nerf: Representing scenes as neural radiance fields for view synthesis.
\newblock \emph{CACM}, 2021.

\bibitem[M{\"u}ller et~al.(2022)M{\"u}ller, Evans, Schied, and Keller]{ngp}
Thomas M{\"u}ller, Alex Evans, Christoph Schied, and Alexander Keller.
\newblock Instant neural graphics primitives with a multiresolution hash encoding.
\newblock \emph{ACM ToG}, 2022.

\bibitem[Ni et~al.(2024)Ni, Zhao, Wang, Zhu, Qin, Huang, Liu, Chen, Wang, Zhang, et~al.]{recondreamer}
Chaojun Ni, Guosheng Zhao, Xiaofeng Wang, Zheng Zhu, Wenkang Qin, Guan Huang, Chen Liu, Yuyin Chen, Yida Wang, Xueyang Zhang, et~al.
\newblock Recondreamer: Crafting world models for driving scene reconstruction via online restoration.
\newblock \emph{arXiv preprint arXiv:2411.19548}, 2024.

\bibitem[Ost et~al.(2021)Ost, Mannan, Thuerey, Knodt, and Heide]{ost2021neural}
Julian Ost, Fahim Mannan, Nils Thuerey, Julian Knodt, and Felix Heide.
\newblock Neural scene graphs for dynamic scenes.
\newblock In \emph{CVPR}, 2021.

\bibitem[Park et~al.(2021)Park, Sinha, Hedman, Barron, Bouaziz, Goldman, Martin-Brualla, and Seitz]{hypernerf}
Keunhong Park, Utkarsh Sinha, Peter Hedman, Jonathan~T Barron, Sofien Bouaziz, Dan~B Goldman, Ricardo Martin-Brualla, and Steven~M Seitz.
\newblock Hypernerf: A higher-dimensional representation for topologically varying neural radiance fields.
\newblock \emph{arXiv preprint arXiv:2106.13228}, 2021.

\bibitem[Podell et~al.(2023)Podell, English, Lacey, Blattmann, Dockhorn, M{\"u}ller, Penna, and Rombach]{sdxl}
Dustin Podell, Zion English, Kyle Lacey, Andreas Blattmann, Tim Dockhorn, Jonas M{\"u}ller, Joe Penna, and Robin Rombach.
\newblock Sdxl: Improving latent diffusion models for high-resolution image synthesis.
\newblock \emph{arXiv preprint arXiv:2307.01952}, 2023.

\bibitem[Rematas et~al.(2022)Rematas, Liu, Srinivasan, Barron, Tagliasacchi, Funkhouser, and Ferrari]{urbannerf}
Konstantinos Rematas, Andrew Liu, Pratul~P Srinivasan, Jonathan~T Barron, Andrea Tagliasacchi, Thomas Funkhouser, and Vittorio Ferrari.
\newblock Urban radiance fields.
\newblock In \emph{CVPR}, 2022.

\bibitem[Rombach et~al.(2022)Rombach, Blattmann, Lorenz, Esser, and Ommer]{sd}
Robin Rombach, Andreas Blattmann, Dominik Lorenz, Patrick Esser, and Bj{\"o}rn Ommer.
\newblock High-resolution image synthesis with latent diffusion models.
\newblock In \emph{CVPR}, 2022.

\bibitem[Song et~al.(2023)Song, Chen, Li, Chen, Chen, Yuan, Xu, and Geiger]{nerfplayer}
Liangchen Song, Anpei Chen, Zhong Li, Zhang Chen, Lele Chen, Junsong Yuan, Yi Xu, and Andreas Geiger.
\newblock Nerfplayer: A streamable dynamic scene representation with decomposed neural radiance fields.
\newblock \emph{IEEE Transactions on Visualization and Computer Graphics}, 2023.

\bibitem[Sun et~al.(2020)Sun, Kretzschmar, Dotiwalla, Chouard, Patnaik, Tsui, Guo, Zhou, Chai, Caine, Vasudevan, Han, Ngiam, Zhao, Timofeev, Ettinger, Krivokon, Gao, Joshi, Zhang, Shlens, Chen, and Anguelov]{waymo}
Pei Sun, Henrik Kretzschmar, Xerxes Dotiwalla, Aurelien Chouard, Vijaysai Patnaik, Paul Tsui, James Guo, Yin Zhou, Yuning Chai, Benjamin Caine, Vijay Vasudevan, Wei Han, Jiquan Ngiam, Hang Zhao, Aleksei Timofeev, Scott Ettinger, Maxim Krivokon, Amy Gao, Aditya Joshi, Yu Zhang, Jonathon Shlens, Zhifeng Chen, and Dragomir Anguelov.
\newblock Scalability in perception for autonomous driving: Waymo open dataset.
\newblock In \emph{CVPR}, 2020.

\bibitem[Tancik et~al.(2022)Tancik, Casser, Yan, Pradhan, Mildenhall, Srinivasan, Barron, and Kretzschmar]{blocknerf}
Matthew Tancik, Vincent Casser, Xinchen Yan, Sabeek Pradhan, Ben Mildenhall, Pratul~P Srinivasan, Jonathan~T Barron, and Henrik Kretzschmar.
\newblock Block-nerf: Scalable large scene neural view synthesis.
\newblock In \emph{CVPR}, 2022.

\bibitem[Tonderski et~al.(2024)Tonderski, Lindstr{\"o}m, Hess, Ljungbergh, Svensson, and Petersson]{Neurad}
Adam Tonderski, Carl Lindstr{\"o}m, Georg Hess, William Ljungbergh, Lennart Svensson, and Christoffer Petersson.
\newblock Neurad: Neural rendering for autonomous driving.
\newblock In \emph{CVPR}, 2024.

\bibitem[Wang et~al.(2024{\natexlab{a}})Wang, Fan, Wang, Chen, and Zhang]{freevs}
Qitai Wang, Lue Fan, Yuqi Wang, Yuntao Chen, and Zhaoxiang Zhang.
\newblock Freevs: Generative view synthesis on free driving trajectory.
\newblock \emph{arXiv preprint arXiv:2410.18079}, 2024{\natexlab{a}}.

\bibitem[Wang et~al.(2023)Wang, Zhu, Huang, Chen, Zhu, and Lu]{drivedreamer}
Xiaofeng Wang, Zheng Zhu, Guan Huang, Xinze Chen, Jiagang Zhu, and Jiwen Lu.
\newblock Drivedreamer: Towards real-world-driven world models for autonomous driving.
\newblock \emph{arXiv preprint arXiv:2309.09777}, 2023.

\bibitem[Wang et~al.(2024{\natexlab{b}})Wang, Zhao, Liu, Wang, Zhao, Bao, Zhu, Zhang, and Wang]{egovid}
Xiaofeng Wang, Kang Zhao, Feng Liu, Jiayu Wang, Guosheng Zhao, Xiaoyi Bao, Zheng Zhu, Yingya Zhang, and Xingang Wang.
\newblock Egovid-5m: A large-scale video-action dataset for egocentric video generation.
\newblock \emph{arXiv preprint arXiv:2411.08380}, 2024{\natexlab{b}}.

\bibitem[Wu et~al.(2024)Wu, Yi, Fang, Xie, Zhang, Wei, Liu, Tian, and Wang]{4dgs}
Guanjun Wu, Taoran Yi, Jiemin Fang, Lingxi Xie, Xiaopeng Zhang, Wei Wei, Wenyu Liu, Qi Tian, and Xinggang Wang.
\newblock 4d gaussian splatting for real-time dynamic scene rendering.
\newblock In \emph{CVPR}, 2024.

\bibitem[Wu et~al.(2023)Wu, Liu, Luo, Zhong, Chen, Xiao, Hou, Lou, Chen, Yang, et~al.]{mars}
Zirui Wu, Tianyu Liu, Liyi Luo, Zhide Zhong, Jianteng Chen, Hongmin Xiao, Chao Hou, Haozhe Lou, Yuantao Chen, Runyi Yang, et~al.
\newblock Mars: An instance-aware, modular and realistic simulator for autonomous driving.
\newblock In \emph{ICAI}, 2023.

\bibitem[Xiao et~al.(2021)Xiao, Shao, Hao, Zhang, Chai, Jiao, Li, Wu, Sun, Jiang, et~al.]{pandaset}
Pengchuan Xiao, Zhenlei Shao, Steven Hao, Zishuo Zhang, Xiaolin Chai, Judy Jiao, Zesong Li, Jian Wu, Kai Sun, Kun Jiang, et~al.
\newblock Pandaset: Advanced sensor suite dataset for autonomous driving.
\newblock In \emph{ITSC}, 2021.

\bibitem[Xie et~al.(2023)Xie, Zhang, Li, Zhang, and Zhang]{snerf}
Ziyang Xie, Junge Zhang, Wenye Li, Feihu Zhang, and Li Zhang.
\newblock S-nerf: Neural radiance fields for street views.
\newblock \emph{arXiv preprint arXiv:2303.00749}, 2023.

\bibitem[Yan et~al.(2024{\natexlab{a}})Yan, Lin, Zhou, Wang, Sun, Zhan, Lang, Zhou, and Peng]{streetgaussian}
Yunzhi Yan, Haotong Lin, Chenxu Zhou, Weijie Wang, Haiyang Sun, Kun Zhan, Xianpeng Lang, Xiaowei Zhou, and Sida Peng.
\newblock Street gaussians for modeling dynamic urban scenes.
\newblock \emph{arXiv preprint arXiv:2401.01339}, 2024{\natexlab{a}}.

\bibitem[Yan et~al.(2024{\natexlab{b}})Yan, Xu, Lin, Jin, Guo, Wang, Zhan, Lang, Bao, Zhou, et~al.]{streetcrafter}
Yunzhi Yan, Zhen Xu, Haotong Lin, Haian Jin, Haoyu Guo, Yida Wang, Kun Zhan, Xianpeng Lang, Hujun Bao, Xiaowei Zhou, et~al.
\newblock Streetcrafter: Street view synthesis with controllable video diffusion models.
\newblock \emph{arXiv preprint arXiv:2412.13188}, 2024{\natexlab{b}}.

\bibitem[Yang et~al.(2023{\natexlab{a}})Yang, Ivanovic, Litany, Weng, Kim, Li, Che, Xu, Fidler, Pavone, et~al.]{emernerf}
Jiawei Yang, Boris Ivanovic, Or Litany, Xinshuo Weng, Seung~Wook Kim, Boyi Li, Tong Che, Danfei Xu, Sanja Fidler, Marco Pavone, et~al.
\newblock Emernerf: Emergent spatial-temporal scene decomposition via self-supervision.
\newblock \emph{arXiv preprint arXiv:2311.02077}, 2023{\natexlab{a}}.

\bibitem[Yang et~al.(2023{\natexlab{b}})Yang, Chen, Wang, Manivasagam, Ma, Yang, and Urtasun]{unisim}
Ze Yang, Yun Chen, Jingkang Wang, Sivabalan Manivasagam, Wei-Chiu Ma, Anqi~Joyce Yang, and Raquel Urtasun.
\newblock Unisim: A neural closed-loop sensor simulator.
\newblock In \emph{CVPR}, 2023{\natexlab{b}}.

\bibitem[Yang et~al.(2024{\natexlab{a}})Yang, Gao, Zhou, Jiao, Zhang, and Jin]{deformablegs}
Ziyi Yang, Xinyu Gao, Wen Zhou, Shaohui Jiao, Yuqing Zhang, and Xiaogang Jin.
\newblock Deformable 3d gaussians for high-fidelity monocular dynamic scene reconstruction.
\newblock In \emph{CVPR}, 2024{\natexlab{a}}.

\bibitem[Yang et~al.(2024{\natexlab{b}})Yang, Pan, Yang, Zhu, and Zhang]{drivex}
Zeyu Yang, Zijie Pan, Yuankun Yang, Xiatian Zhu, and Li Zhang.
\newblock Driving scene synthesis on free-form trajectories with generative prior.
\newblock \emph{arXiv preprint arXiv:2412.01717}, 2024{\natexlab{b}}.

\bibitem[Yang et~al.(2024{\natexlab{c}})Yang, Teng, Zheng, Ding, Huang, Xu, Yang, Hong, Zhang, Feng, et~al.]{yang2024cogvideox}
Zhuoyi Yang, Jiayan Teng, Wendi Zheng, Ming Ding, Shiyu Huang, Jiazheng Xu, Yuanming Yang, Wenyi Hong, Xiaohan Zhang, Guanyu Feng, et~al.
\newblock Cogvideox: Text-to-video diffusion models with an expert transformer.
\newblock \emph{arXiv preprint arXiv:2408.06072}, 2024{\natexlab{c}}.

\bibitem[Yu et~al.(2024{\natexlab{a}})Yu, Chen, Huang, Sattler, and Geiger]{mipgs}
Zehao Yu, Anpei Chen, Binbin Huang, Torsten Sattler, and Andreas Geiger.
\newblock Mip-splatting: Alias-free 3d gaussian splatting.
\newblock In \emph{CVPR}, 2024{\natexlab{a}}.

\bibitem[Yu et~al.(2024{\natexlab{b}})Yu, Wang, Yang, Wang, Xie, Cai, Cao, Ji, and Sun]{sgd}
Zhongrui Yu, Haoran Wang, Jinze Yang, Hanzhang Wang, Zeke Xie, Yunfeng Cai, Jiale Cao, Zhong Ji, and Mingming Sun.
\newblock Sgd: Street view synthesis with gaussian splatting and diffusion prior.
\newblock \emph{arXiv preprint arXiv:2403.20079}, 2024{\natexlab{b}}.

\bibitem[Zermas et~al.(2017)Zermas, Izzat, and Papanikolopoulos]{zermas2017fast}
Dimitris Zermas, Izzat Izzat, and Nikolaos Papanikolopoulos.
\newblock Fast segmentation of 3d point clouds: A paradigm on lidar data for autonomous vehicle applications.
\newblock In \emph{ICRA}, 2017.

\bibitem[Zhang et~al.(2018)Zhang, Isola, Efros, Shechtman, and Wang]{lpips}
Richard Zhang, Phillip Isola, Alexei~A Efros, Eli Shechtman, and Oliver Wang.
\newblock The unreasonable effectiveness of deep features as a perceptual metric.
\newblock In \emph{CVPR}, 2018.

\bibitem[Zhao et~al.(2024{\natexlab{a}})Zhao, Ni, Wang, Zhu, Huang, Chen, Wang, Zhang, Mei, and Wang]{drivedreamer4d}
Guosheng Zhao, Chaojun Ni, Xiaofeng Wang, Zheng Zhu, Guan Huang, Xinze Chen, Boyuan Wang, Youyi Zhang, Wenjun Mei, and Xingang Wang.
\newblock Drivedreamer4d: World models are effective data machines for 4d driving scene representation.
\newblock \emph{arXiv preprint arXiv:2410.13571}, 2024{\natexlab{a}}.

\bibitem[Zhao et~al.(2024{\natexlab{b}})Zhao, Wang, Zhu, Chen, Huang, Bao, and Wang]{drivedreamer2}
Guosheng Zhao, Xiaofeng Wang, Zheng Zhu, Xinze Chen, Guan Huang, Xiaoyi Bao, and Xingang Wang.
\newblock Drivedreamer-2: Llm-enhanced world models for diverse driving video generation.
\newblock \emph{arXiv preprint arXiv:2403.06845}, 2024{\natexlab{b}}.

\bibitem[Zhou et~al.(2024{\natexlab{a}})Zhou, Lin, Wang, Lu, Bai, Liu, Wang, Geiger, and Liao]{zhou2024hugsim}
Hongyu Zhou, Longzhong Lin, Jiabao Wang, Yichong Lu, Dongfeng Bai, Bingbing Liu, Yue Wang, Andreas Geiger, and Yiyi Liao.
\newblock Hugsim: A real-time, photo-realistic and closed-loop simulator for autonomous driving.
\newblock \emph{arXiv preprint arXiv:2412.01718}, 2024{\natexlab{a}}.

\bibitem[Zhou et~al.(2024{\natexlab{b}})Zhou, Lin, Shan, Wang, Sun, and Yang]{drivinggaussian}
Xiaoyu Zhou, Zhiwei Lin, Xiaojun Shan, Yongtao Wang, Deqing Sun, and Ming-Hsuan Yang.
\newblock Drivinggaussian: Composite gaussian splatting for surrounding dynamic autonomous driving scenes.
\newblock In \emph{CVPR}, pages 21634--21643, 2024{\natexlab{b}}.

\bibitem[Zhu et~al.(2024)Zhu, Wang, Zhao, Min, Deng, Dou, Wang, Shi, Wang, Zhang, et~al.]{zhu2024sora}
Zheng Zhu, Xiaofeng Wang, Wangbo Zhao, Chen Min, Nianchen Deng, Min Dou, Yuqi Wang, Botian Shi, Kai Wang, Chi Zhang, et~al.
\newblock Is sora a world simulator? a comprehensive survey on general world models and beyond.
\newblock \emph{arXiv preprint arXiv:2405.03520}, 2024.

\end{thebibliography}
}

\newpage
\twocolumn[{%
\maketitle
\begin{center}
\centering
\resizebox{\linewidth}{!}{
\begin{tabular}{lcc}
    \toprule
    \makecell[c]{Scene} & Start Frame & End Frame \\ \midrule
segment-10359308928573410754\_720\_000\_740\_000\_with\_camera\_labels.tfrecord & 120 & 159 \\
segment-11450298750351730790\_1431\_750\_1451\_750\_with\_camera\_labels.tfrecord & 0 & 39 \\
segment-12496433400137459534\_120\_000\_140\_000\_with\_camera\_labels.tfrecord & 110 & 149 \\
segment-15021599536622641101\_556\_150\_576\_150\_with\_camera\_labels.tfrecord & 0 & 39 \\
segment-16767575238225610271\_5185\_000\_5205\_000\_with\_camera\_labels.tfrecord & 0 & 39 \\
segment-17860546506509760757\_6040\_000\_6060\_000\_with\_camera\_labels.tfrecord & 90 & 129 \\
segment-3015436519694987712\_1300\_000\_1320\_000\_with\_camera\_labels.tfrecord & 40 & 79 \\
segment-6637600600814023975\_2235\_000\_2255\_000\_with\_camera\_labels.tfrecord & 70 & 109 \\
\bottomrule
\end{tabular}
}
\captionof{table}{Eight scenes from the Waymo dataset \cite{waymo} featuring high interactive activity, numerous vehicles, and complex driving trajectories.}
\label{tab:waymo_scene}
\end{center}
}]

In the supplementary material, we provide further details about the datasets and an explanation of the evaluation metrics. Additionally, we present more extensive qualitative results conducted on these datasets.

\section{Datasets and Evaluation Metrics}

\noindent \textbf{Waymo.} For the selection of Waymo \cite{waymo} scenes, we follow the approach used in ReconDreamer \cite{recondreamer}. These eight scenes are characterized by their rich dynamic foregrounds and large-scale variations. The specific scene names are listed in Tab.~\ref{tab:waymo_scene}.

\noindent \textbf{nuScenes.} Similarly, we select eight scenes from the nuScenes dataset \cite{nuscenes}, each featuring complex traffic environments with diverse and challenging conditions. The specific scenes are as follows: \textit{0037}, \textit{0040}, \textit{0050}, \textit{0062}, \textit{0064}, \textit{0086}, \textit{0087}, \textit{0100}.

\noindent \textbf{PandaSet.} The selection of scenes from the PandaSet dataset \cite{pandaset} follows the approach used in UniSim \cite{unisim} and NeuRAD \cite{Neurad}. The specific scenes are \textit{001}, \textit{011}, \textit{016}, \textit{028}, \textit{053}, \textit{063}, \textit{084}, \textit{106}, \textit{123}, and \textit{158}.

\noindent \textbf{EUVS.} EUVS \cite{euvs} is a dataset specifically designed for extrapolated urban views, where all results across different lanes are captured from real-world observations. However, since the data for different lanes are not collected simultaneously, the dataset primarily focuses on static backgrounds, requiring the foreground to be ignored during evaluation. Consequently, when calculating metrics such as PSNR, SSIM, and LPIPS, the foreground must first be masked out to ensure accurate assessment of the static scene reconstruction. The four scenes from Level 1 are: \textit{vegas\_location\_1}, \textit{vegas\_location\_2}, \textit{vegas\_location\_15}, \textit{vegas\_location\_22}. 

\begin{figure*}[t] 
\centering
\includegraphics[width=0.98\textwidth]{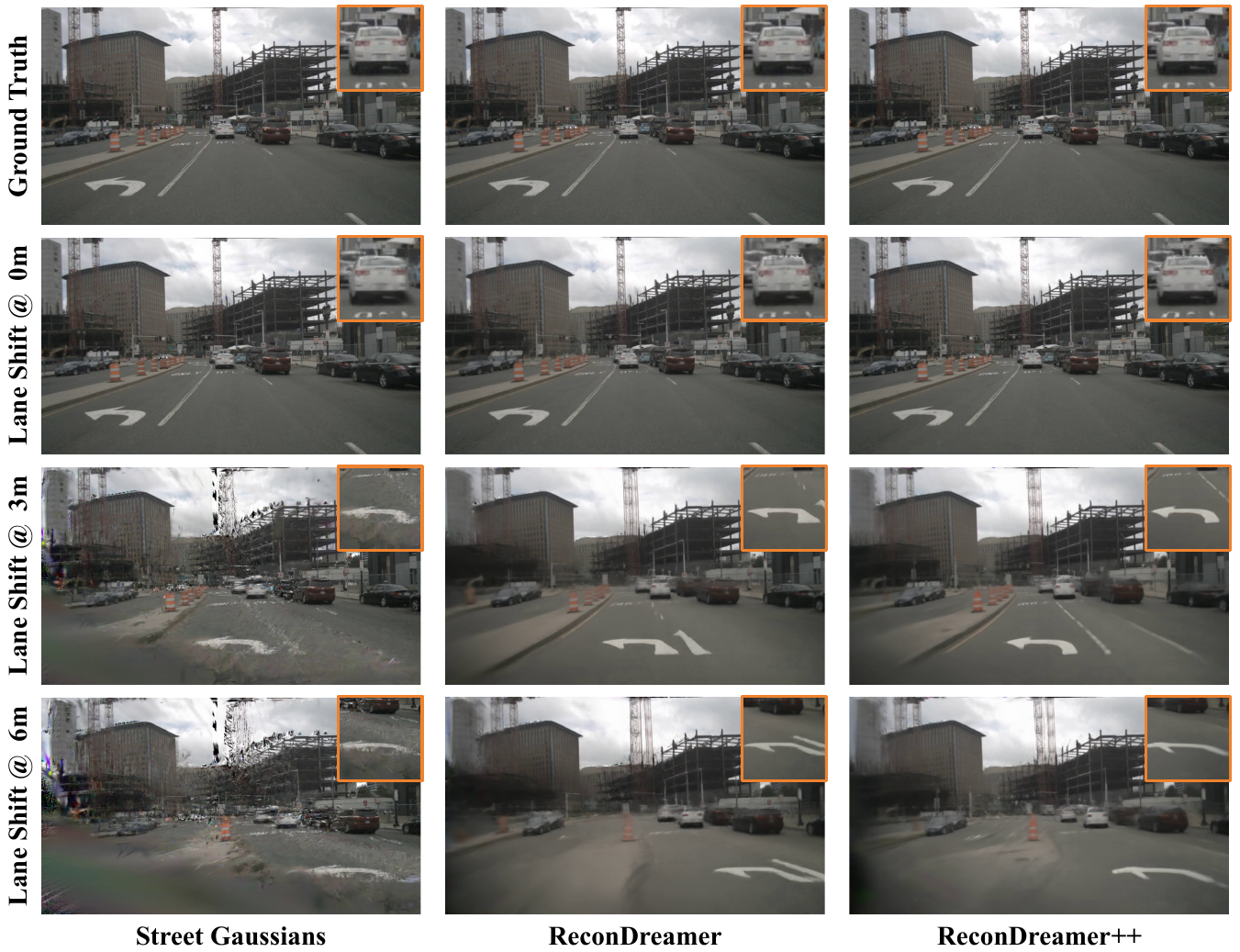}
\caption{Qualitative comparisons of different trajectory renderings on nuScenes \cite{nuscenes}. The orange boxes highlight that ReconDreamer++ significantly enhances the rendering quality with Street Gaussians \cite{streetgaussian} and ReconDreamer \cite{recondreamer}.}
\label{fig_nuscenes}
\end{figure*}

\begin{figure*}[t] 
\centering
\includegraphics[width=\textwidth]{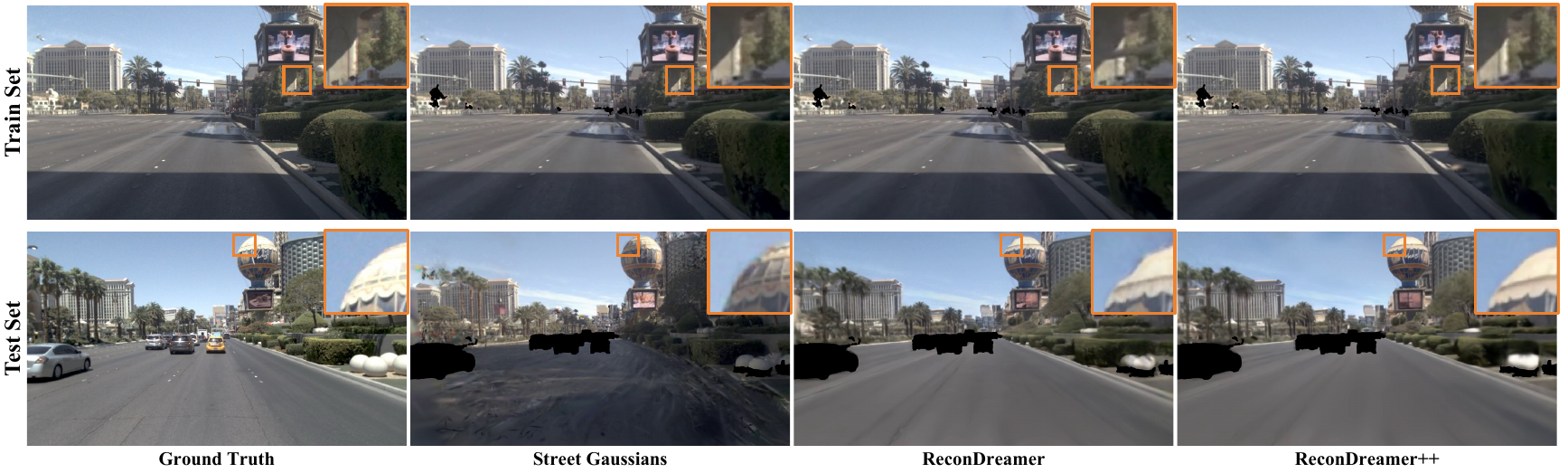}
\caption{Qualitative comparisons of different trajectory renderings on EUVS \cite{euvs}. The orange boxes highlight that ReconDreamer++ significantly enhances the rendering quality with Street Gaussians \cite{streetgaussian} and ReconDreamer \cite{recondreamer}.}
\label{fig_euvs}
\end{figure*}

\begin{figure*}[t] 
\centering
\includegraphics[width=0.99\textwidth]{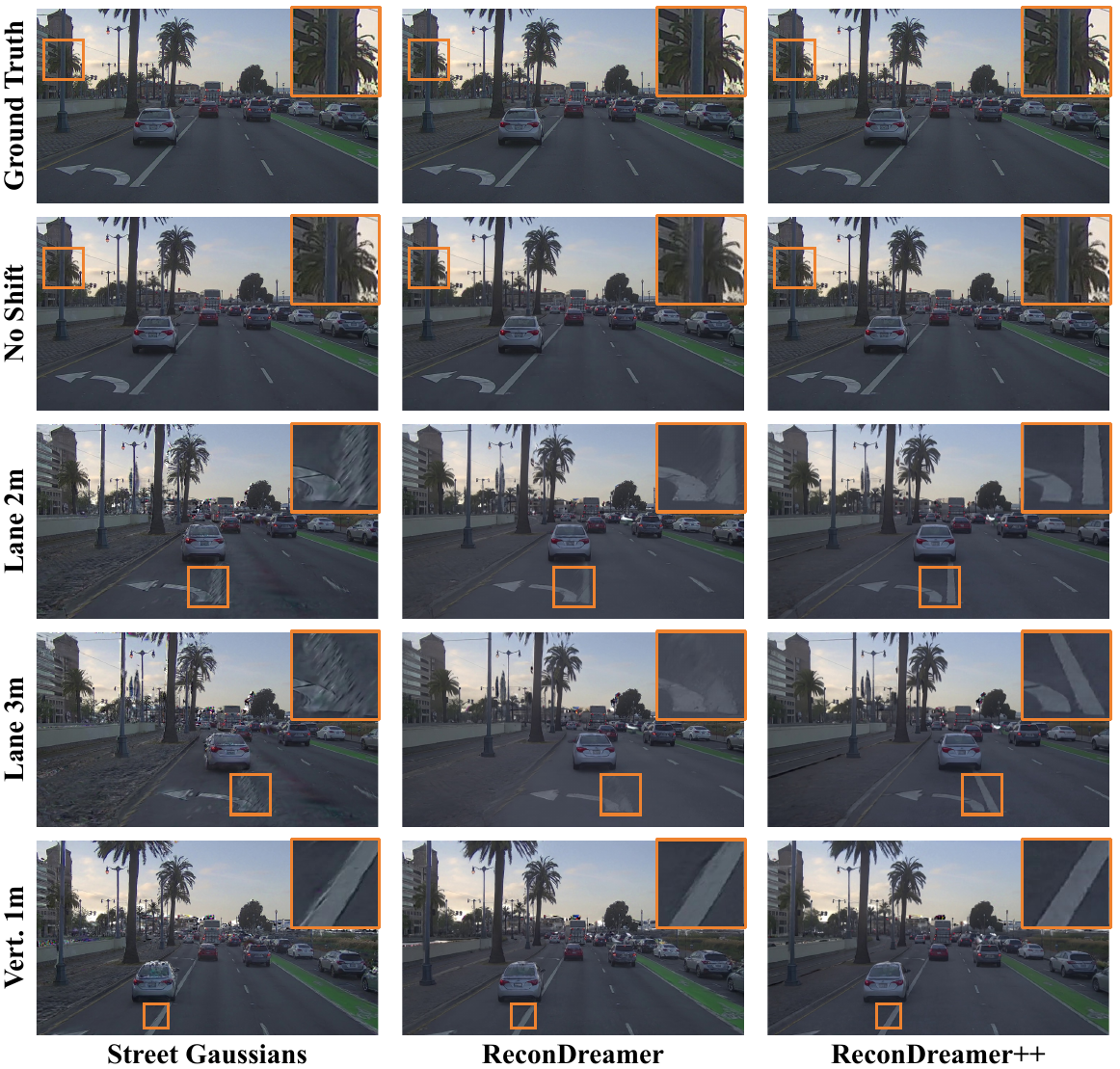}
\caption{Qualitative comparisons of different trajectory renderings on PandaSet \cite{pandaset}. The orange boxes highlight that ReconDreamer++ significantly enhances the rendering quality with Street Gaussians \cite{streetgaussian} and ReconDreamer \cite{recondreamer}.}
\label{fig_pandaset}
\end{figure*}

\noindent \textbf{Metrics.} As mentioned in the main text, we utilize Novel Trajectory Agent Intersection over Union (NTA-IoU) and Novel Trajectory Lane Intersection over Union (NTL-IoU) to assess the quality of the rendered video, both metrics proposed in DriveDreamer4D \cite{drivedreamer4d}. These metrics are specifically designed to evaluate the spatiotemporal coherence of foreground agents and background lanes, respectively.

The NTA-IoU processes images rendered under new trajectories using the YOLO11 \cite{yolo11_ultralytics} detector to extract 2D bounding boxes of vehicles. Meanwhile, by applying geometric transformations to the 3D bounding boxes from the original trajectories, they can be accurately projected onto the new trajectory perspective, thus obtaining the ground truth 2D bounding boxes in the new trajectory view. Each projected 2D bounding box will find the nearest 2D bounding box generated by the detector and compute their Intersection over Union (IoU). 

Similarly, the NTL-IoU employs the TwinLiteNet \cite{che2023twinlitenet} model to detect lane in the images rendered under the new trajectories, and the lane from the original trajectories will also be projected onto the new trajectory through corresponding geometric transformations.  Finally, the mean Intersection over Union (IoU) between the projected and detected lane lines is calculated.

\section{More Qualitative Results}

The visualization and comparison results for all datasets can be found in the \textit{./video\_comparison} folder. This folder contains detailed comparisons for each dataset under both original and novel trajectories, along with ablation study videos that highlight the contributions and impact of different components.

\noindent
\textbf{nuScenes.} The visualization results for the nuScenes dataset are presented in Fig.~\ref{fig_nuscenes}, offering a comprehensive comparison of the performance of different methods. The experimental findings demonstrate that ReconDreamer++ achieves results on the original trajectory that are comparable to traditional reconstruction techniques such as Street Gaussians \cite{streetgaussian}. Notably, ReconDreamer++ even outperforms Street Gaussians in certain fine-grained details, showcasing its ability to capture intricate structures within the scene. In contrast, ReconDreamer \cite{recondreamer} exhibits limitations, producing relatively blurry and less detailed rendering results on the original trajectory, which highlights the challenges faced by earlier approaches. When evaluating novel trajectories, the visualizations reveal the superior capabilities of ReconDreamer++. Specifically, the method demonstrates exceptional performance in maintaining high geometric consistency of structured elements, particularly the ground surface, with respect to the ground truth. As highlighted by the orange bounding boxes in Fig.~\ref{fig_nuscenes}, ReconDreamer++ accurately renders left-turn arrows on the ground under novel trajectories, maintaining strong geometric fidelity. In contrast, both ReconDreamer and Street Gaussians struggle to render these ground structures with sufficient accuracy, often leading to distorted or incomplete representations. Such inaccuracies can significantly impact downstream tasks in autonomous driving, where precise understanding of road markings and other structured elements is essential for safe navigation.

\noindent
\textbf{PandaSet.} The visualization results for PandaSet \cite{pandaset} are shown in Fig.~\ref{fig_pandaset}, with comparisons that align with the experimental findings on Waymo \cite{waymo} and nuScenes \cite{nuscenes}. ReconDreamer++ demonstrates a comprehensive improvement over ReconDreamer \cite{recondreamer} in rendering quality for both original and novel trajectories. Notably, it not only ensures rendering performance on par with traditional reconstruction methods, such as Street Gaussians \cite{streetgaussian}, for original trajectories, but also achieves state-of-the-art (SOTA) results for novel trajectories. In particular, ReconDreamer++ excels in rendering structured elements such as lane markings on the ground, achieving exceptional accuracy and consistency. As highlighted by the orange bounding boxes in Fig.~\ref{fig_pandaset}, the method accurately reconstructs these critical road features even under challenging novel viewpoints. These results further validate the effectiveness and robustness of ReconDreamer++ across diverse datasets and scenarios.

\noindent
\textbf{EUVS.} The experimental results for EUVS \cite{euvs} are shown in Fig.~\ref{fig_euvs}, demonstrating the superior performance of ReconDreamer++ on both the training set and the test set. Since this dataset consists of data collected from the same scene at different times, the dynamic object regions—including cars, pedestrians, and other moving elements—in the rendered results are masked to facilitate a clearer and more accurate comparison. Specifically, as highlighted by the orange bounding boxes in Fig.~\ref{fig_euvs}, ReconDreamer++ maintains an exceptionally high level of consistency with the ground truth in terms of detail rendering.

\end{document}